\documentclass{article}
\usepackage[preprint]{neurips_2025}

\usepackage[utf8]{inputenc}
\usepackage[T1]{fontenc}
\usepackage{hyperref}
\usepackage{url}
\usepackage{booktabs}
\usepackage{amsmath, amssymb}
\usepackage{microtype}
\usepackage{xcolor}
\usepackage{enumitem}
\usepackage{graphicx}

\title{Mask-Conditioned Voxel Diffusion for Joint Geometry and Color Inpainting}

\author{
  Aarya Sumuk \\
  Department of Mechanical Engineering \\
  Stanford University \\
  \texttt{asumuk@stanford.edu} \\
}

\begin{document}

\maketitle

\begin{abstract}
We present a lightweight two-stage framework for joint geometry and color inpainting of damaged 3D objects, motivated by the digital restoration of cultural heritage artifacts. The pipeline separates damage localization from reconstruction. In the first stage, a 2D convolutional network predicts damage masks on RGB slices extracted from a voxelized object, and these predictions are aggregated into a volumetric mask. In the second stage, a diffusion-based 3D U-Net performs mask-conditioned inpainting directly on voxel grids, reconstructing geometry and color while preserving observed regions. The model jointly predicts occupancy and color using a composite objective that combines occupancy reconstruction with masked color reconstruction and perceptual regularization. We evaluate the approach on a curated set of textured artifacts with synthetically generated damage using standard geometric and color metrics. Compared to symmetry-based baselines, our method produces more complete geometry and more coherent color reconstructions at a fixed $32^3$ resolution. Overall, the results indicate that explicit mask conditioning is a practical way to guide volumetric diffusion models for joint 3D geometry and color inpainting.
\end{abstract}

%--------------------------------------
% Introduction
%--------------------------------------
\section{Introduction}

Reconstructing incomplete or damaged 3D objects is a long-standing problem in computer vision and graphics, with applications spanning cultural heritage preservation, robotics, and general shape understanding. In the context of heritage artifacts, 3D scans of sculptures, ceramics, and small objects frequently contain missing regions due to erosion, breakage, or limitations of the acquisition process. Digital restoration of such scans enables noninvasive study, visualization, and archival of fragile objects, while avoiding physical interventions that may introduce irreversible changes.

From a technical perspective, this problem can be framed as joint completion of 3D geometry and appearance given a partial observation. While classical reconstruction methods focus primarily on geometry, many modern scans include color or texture information that can provide additional context for restoration. At the same time, methods that operate on high-resolution meshes or implicit representations often require substantial computational resources, which can limit their practicality in lightweight or exploratory settings.

In this work, we study joint geometry and color inpainting in a voxel-based representation. Starting from a damaged textured mesh, we voxelize the object into a binary occupancy grid together with an aligned per-voxel RGB volume. Missing regions are synthetically generated by removing structured portions of the volume using slice-wise hole masks and morphological erosion, yielding paired damaged and intact examples. The objective is to reconstruct both a completed occupancy grid and a corresponding color volume that approximate the original undamaged shape and appearance.

To address this problem, we adopt a two-stage pipeline that decouples damage localization from content reconstruction. In the first stage, a 2D convolutional U-Net predicts damage masks on individual RGB slices extracted from the voxel grid. Slice-level predictions are aggregated to produce a volumetric mask that identifies regions requiring inpainting. This explicit localization step simplifies the subsequent reconstruction task by allowing the model to focus on filling missing regions rather than detecting them.

In the second stage, we perform 3D inpainting using a diffusion-based U-Net operating directly on voxel grids. The network takes as input the masked occupancy, the predicted damage mask, and the masked color channels, and is trained to reconstruct missing geometry and color while preserving intact regions. Diffusion timesteps and sinusoidal time embeddings are used to guide the denoising process, following recent work on diffusion models for 3D shape generation and completion \cite{nam2022_3dldm,schroppel2024_scdiff}. The training objective combines binary cross-entropy for occupancy reconstruction with masked $L_1$ loss for color, along with optional perceptual regularization computed on 2D slices and a weak color prior tailored to ceramic-like materials.

We evaluate the proposed approach on a small curated dataset of textured artifacts, consisting of publicly available 3D scans and CAD models. Performance is measured using standard geometric metrics, including Chamfer distance and F-score at a 1~mm threshold, as well as masked MSE and PSNR for color fidelity. Comparisons are made against simple baselines such as symmetry-based inpainting and Poisson surface reconstruction followed by voxelization \cite{kazhdan2006poisson,kazhdan2013screened}. Across held-out examples, the diffusion-based model produces more complete geometry and more coherent color reconstructions than these baselines, while remaining computationally lightweight.

While the scope of this study is limited to moderate voxel resolutions and synthetically generated damage, the results suggest that explicitly conditioning 3D diffusion models on predicted damage masks is a viable strategy for joint geometry and color inpainting under constrained settings.

%--------------------------------------
% Related Work
%--------------------------------------
\section{Related Work}

Our work is related to prior research on geometric reconstruction, volumetric shape completion, diffusion-based generative modeling, and appearance inpainting. We briefly review these areas and clarify how our approach fits within existing literature.

\subsection{Geometric Reconstruction and Completion}

Classical approaches to repairing incomplete 3D geometry rely on surface interpolation and implicit reconstruction. Poisson surface reconstruction and its screened variant formulate surface completion as the solution to a Poisson equation defined over oriented point clouds, producing watertight meshes from partial data \cite{kazhdan2006poisson,kazhdan2013screened}. While effective for smooth surfaces, these methods tend to oversmooth high-frequency detail and do not explicitly reason about the location or structure of missing regions.

Learning-based approaches address shape completion using volumetric or point-based representations. Early voxel-based models reconstruct shapes from one or more views using encoder--decoder architectures \cite{choy2016r2n2}, while later work extends these ideas to scene-scale completion tasks \cite{dai2018scancomplete}. Point-cloud methods infer missing geometry directly from unordered point sets \cite{yuan2018pcn,yang2018foldingnet}, but typically require additional surface reconstruction steps to obtain watertight meshes.

\subsection{Slice-Based and 2.5D Processing}

Processing volumetric data via 2D slices has been widely adopted in medical imaging as a compromise between computational efficiency and spatial context. In so-called 2.5D approaches, independent 2D predictions from multiple orthogonal views are aggregated to form a 3D result, achieving competitive performance without the cost of full 3D convolutions \cite{cicek2016unet3d}.

Our first stage adopts a similar strategy by predicting damage masks on 2D RGB slices extracted from a voxelized object and aggregating them into a volumetric mask. This explicit localization of missing regions simplifies subsequent 3D reconstruction and allows the diffusion model to focus on inpainting rather than detection.

\subsection{Diffusion Models for 3D Shape Completion}

Diffusion models have recently emerged as a powerful class of generative models for 3D data. Several works perform diffusion in latent spaces learned from implicit shape representations, enabling high-resolution shape synthesis at the cost of substantial training complexity \cite{nam2022_3dldm}. Other approaches apply diffusion directly to voxel grids or occupancy fields, achieving strong completion performance on CAD-style datasets \cite{schroppel2024_scdiff}.

In contrast to these methods, our approach applies diffusion at a relatively low voxel resolution and conditions explicitly on a predicted damage mask. This design prioritizes computational simplicity and controllability over high-resolution synthesis.

\subsection{Color and Appearance Inpainting}

Compared to geometric completion, appearance inpainting for 3D data has received comparatively less attention. Prior work has explored predicting colors or textures for meshes using perceptual losses, often treating geometry as fixed and focusing on appearance transfer rather than joint reconstruction \cite{johnson2016perceptual}. 

Our method performs joint geometry and color inpainting within a unified voxel-based diffusion framework. Color is predicted as a residual conditioned on masked geometry and intact appearance, and is regularized using masked reconstruction losses and perceptual features computed on 2D slices using pretrained convolutional networks \cite{simonyan2014vgg}.

%--------------------------------------
% Methods
%--------------------------------------
\section{Methods}
\label{sec:methods}

We propose a two-stage pipeline for joint geometry and color inpainting of damaged 3D objects using a voxel-based representation.
All experiments are conducted at a fixed spatial resolution of $32\times32\times32$, enabling efficient training and evaluation.
The first stage estimates volumetric damage masks via slice-wise 2D segmentation, and the second stage performs mask-conditioned 3D inpainting using a diffusion-based U-Net.

\subsection{Voxel Representation}

Each artifact is represented as a binary occupancy grid
\[
V^{\mathrm{dam}} \in \{0,1\}^{32\times32\times32},
\]
together with an aligned RGB color volume
\[
C^{\mathrm{dam}} \in [0,1]^{3\times32\times32\times32}.
\]
Voxelization is performed after normalizing meshes to a unit cube.
Color values are assigned to occupied voxels via texture rendering, while empty voxels are assigned zero color.

Synthetic damage is introduced by removing structured regions from the voxel grid using slice-wise hole masks and morphological erosion.
This process yields paired damaged volumes $(V^{\mathrm{dam}}, C^{\mathrm{dam}})$ and corresponding ground-truth intact volumes $(V^{\mathrm{gt}}, C^{\mathrm{gt}})$, as well as a binary damage mask
\[
M = \mathbb{1}[V^{\mathrm{gt}} = 1 \wedge V^{\mathrm{dam}} = 0].
\]

\subsection{Stage 1: Damage Mask Prediction}

The first stage predicts a refined volumetric damage mask from the damaged color volume.
We operate on axial RGB slices extracted from $C^{\mathrm{dam}}$.
For each slice index $z$, a 4-channel input is formed by concatenating the RGB slice with a coarse binary damage indicator computed from the damaged volume generation procedure:
\[
x_z \in \mathbb{R}^{4\times32\times32}.
\]

A 2D U-Net predicts a per-pixel probability map $\hat{M}_z \in [0,1]^{32\times32}$, which is thresholded at 0.5 to obtain a binary slice-level mask.
Predictions across all slices are stacked along the depth dimension and combined using a logical OR operation.
A small 3D morphological closing operation with a $3\times3\times3$ structuring element is applied to remove isolated gaps, producing the final volumetric mask
\[
\widehat{M} \in \{0,1\}^{32\times32\times32}.
\]

\paragraph{Architecture and training.}
The 2D mask network follows a standard U-Net architecture with four downsampling stages (64–512 channels) and symmetric upsampling with skip connections.
The network is trained for 50 epochs using binary cross-entropy loss and the Adam optimizer with learning rate $10^{-4}$.
Data augmentation includes random flips and in-plane rotations up to $\pm15^\circ$.

\subsection{Stage 2: Mask-Conditioned Diffusion Inpainting}

The second stage reconstructs missing geometry and color conditioned on the predicted damage mask.
Masked inputs are computed as
\[
V^{\mathrm{mask}} = V^{\mathrm{dam}} \odot (1 - \widehat{M}), \qquad
C^{\mathrm{mask}} = C^{\mathrm{dam}} \odot (1 - \widehat{M}),
\]
where $\odot$ denotes elementwise multiplication.

The diffusion model input is a 5-channel voxel tensor
\[
X = [\,V^{\mathrm{mask}} \;\Vert\; \widehat{M} \;\Vert\; C^{\mathrm{mask}}\,]
\in \mathbb{R}^{5\times32\times32\times32}.
\]

We employ a denoising diffusion probabilistic model with a linear noise schedule over $T=1000$ timesteps.
At each training iteration, a timestep $t$ is sampled uniformly and encoded using a sinusoidal embedding injected at the network bottleneck.

\paragraph{Network architecture.}
The diffusion backbone is a 3D U-Net with four resolution levels.
Each encoder block consists of two $3\times3\times3$ convolutions with ReLU activations and residual connections, followed by $2\times2\times2$ max pooling.
Channel width increases from 32 to 128 across levels.
The decoder mirrors this structure using transpose convolutions and skip connections.
Two separate $1\times1\times1$ heads predict an occupancy logit volume and a 3-channel color residual.

Final occupancy is obtained by applying a sigmoid and thresholding at 0.5.
Final color values are computed by adding the predicted residual to $C^{\mathrm{mask}}$ within masked regions only.

\paragraph{Training objective.}
The model is trained using a weighted combination of losses:
(i) a diffusion noise-prediction loss (MSE between predicted and target noise on the occupancy channel),
(ii) binary cross-entropy on final occupancy probabilities,
(iii) masked $L_1$ loss on RGB values inside $\widehat{M}$,
(iv) slice-wise perceptual loss computed using VGG-16 features up to \texttt{conv3\_3}, and
(v) a weak color prior encouraging plausible ceramic-like color distributions.
Loss weights are set to 1.0 for occupancy BCE, 20.0 for color $L_1$, and 0.1 for perceptual and color prior terms.

\paragraph{Optimization.}
Training is performed for 100 epochs using the Adam optimizer with initial learning rate $10^{-3}$ and batch size 4 volumes.
A ReduceLROnPlateau scheduler with factor 0.5 and patience 5 is used.
Random mirroring along the axial direction is applied with probability 0.5.

%--------------------------------------
% Dataset and Preprocessing
%--------------------------------------
\section{Dataset and Preprocessing}
\label{sec:data}

\subsection{Artifact Collection}

We curated a dataset of 23 porcelain-style artifacts composed of both real-world scans and CAD models.
Specifically, the dataset includes 9 textured 3D scans from the Smithsonian 3D Scan Collection (public domain) and 14 CAD models obtained from Free3D under Creative Commons licenses.
These objects were selected to cover a range of shape complexities, including both smooth, symmetric forms and more intricate decorative geometries.

All meshes were manually inspected to remove disconnected components and were decimated to approximately 100{,}000 faces using quadric edge collapse decimation in MeshLab.
The dataset was split at the object level into 16 training artifacts, 4 validation artifacts, and 3 test artifacts, ensuring that shape complexity was balanced across splits.

\begin{figure}[t]
  \centering
  \includegraphics[width=1.0\linewidth]{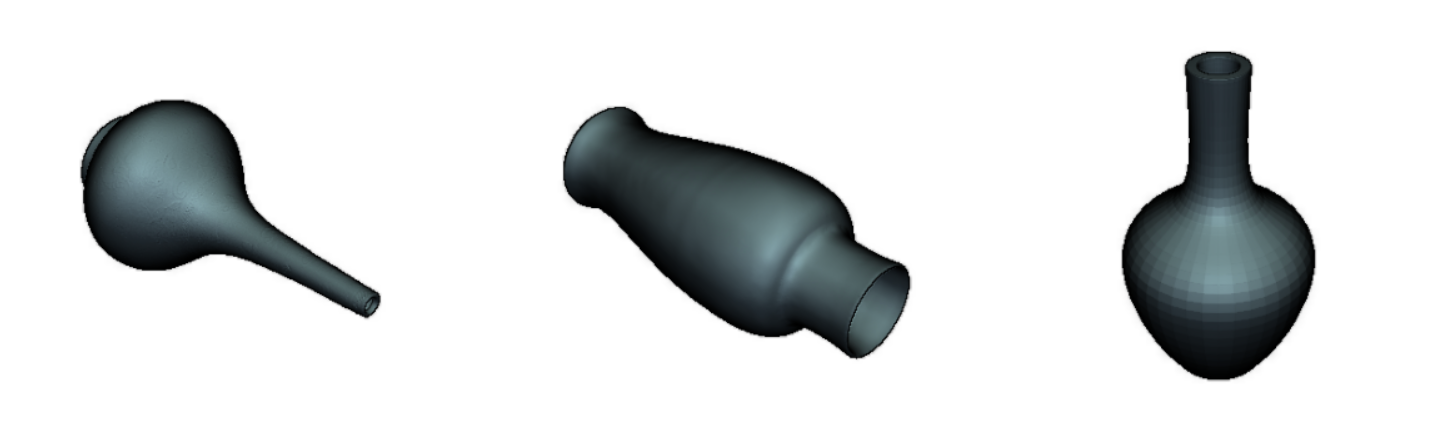}
  \caption{Representative meshes from the curated artifact dataset.}
  \label{fig:gallery}
\end{figure}

\subsection{Voxelization and Color Volumes}

Each mesh is normalized to fit within a unit cube centered at the origin and voxelized at a fixed resolution of $32\times32\times32$.
Voxelization is performed using a ray-casting approach, where voxels whose centers intersect the mesh surface are marked as occupied. Although the original meshes are specified in millimeters, all objects are uniformly normalized prior to voxelization; as a result, reported distances correspond to a consistent relative scale, and we treat one voxel as one millimeter for evaluation purposes.

A flood-fill from the exterior is then applied to produce a watertight binary occupancy grid
\[
V^{\mathrm{gt}} \in \{0,1\}^{32\times32\times32}.
\]

Per-voxel RGB color values are rendered from the original textured meshes using Blender and aligned with the voxel grid, yielding a color volume
\[
C^{\mathrm{gt}} \in [0,1]^{3\times32\times32\times32}.
\]
Color values are defined only for occupied voxels; unoccupied voxels are assigned zero color.

\begin{figure}[t]
  \centering
  \includegraphics[width=0.45\linewidth]{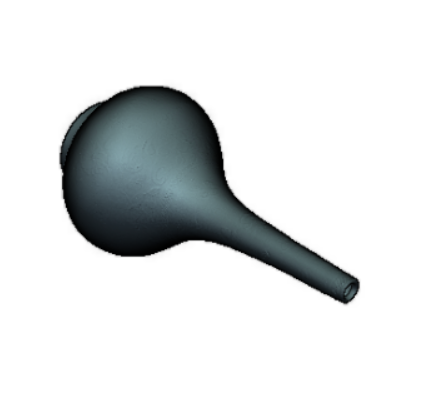}
  \hfill
  \includegraphics[width=0.45\linewidth]{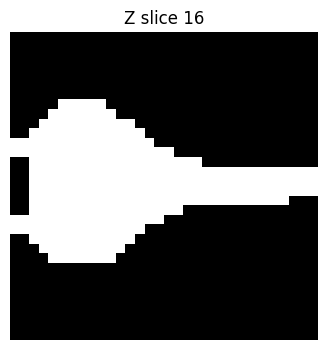}
  \caption{Example mesh and corresponding $32^3$ voxel occupancy grid.}
  \label{fig:voxelization}
\end{figure}

\subsection{Synthetic Damage Generation}

To generate training pairs for inpainting, synthetic damage is applied directly to the voxelized ground-truth volumes.
For each axial slice, between one and three random hole regions are sampled.
These regions are either circular or polygonal, with radii uniformly drawn from 5 to 10 voxels, and are removed from the occupancy grid on the corresponding slice.

After slice-wise hole removal, a 3D binary erosion with a spherical structuring element of radius two voxels is applied to the entire volume to simulate surface abrasion and chipping.
This process yields a damaged occupancy grid
\[
V^{\mathrm{dam}} \in \{0,1\}^{32\times32\times32},
\]
along with a binary damage mask
\[
M = \mathbb{1}\!\left[V^{\mathrm{gt}} = 1 \;\wedge\; V^{\mathrm{dam}} = 0\right].
\]

The damaged color volume is computed by masking the ground-truth color volume:
\[
C^{\mathrm{dam}} = C^{\mathrm{gt}} \odot V^{\mathrm{dam}},
\]
where $\odot$ denotes elementwise multiplication.

\begin{figure}[t]
  \centering
  \begin{tabular}{@{}c@{\;}c@{\;}c@{}}
    \includegraphics[width=0.3\linewidth]{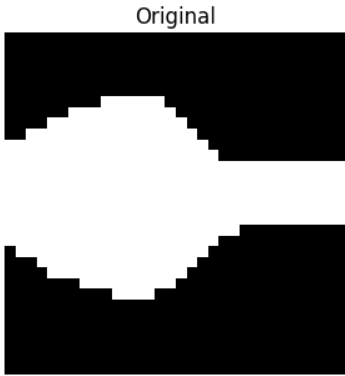} &
    \includegraphics[width=0.3\linewidth]{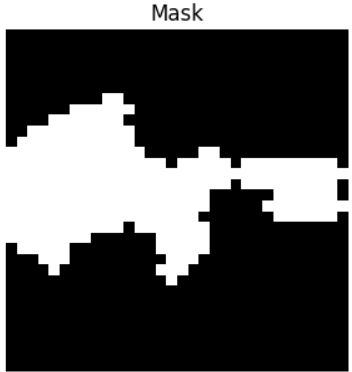} &
    \includegraphics[width=0.3\linewidth]{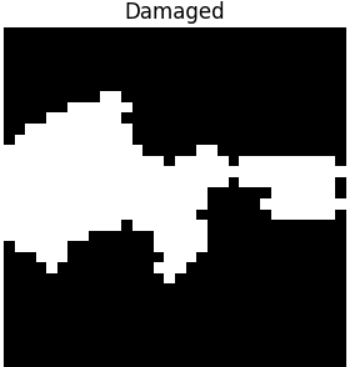} \\
    (a) Intact slice & (b) Damage mask & (c) Damaged slice
  \end{tabular}
  \caption{Example axial voxel slices showing intact geometry, synthetic damage mask, and damaged input.}
  \label{fig:damage_slices}
\end{figure}

\subsection{Normalization and Data Augmentation}

All RGB values are normalized to the range $[0,1]$.
During Stage~1 mask prediction training, RGB slices are standardized using ImageNet mean and standard deviation, consistent with the pretrained encoder initialization.
For Stage~2 diffusion training, voxel occupancy and RGB values are used directly in $[0,1]$ without further normalization.

Data augmentation for Stage~1 includes random horizontal and vertical flips and in-plane rotations up to $\pm15^\circ$.
For Stage~2, random mirroring along the axial direction is applied with probability 0.5 to encourage symmetry-aware generalization.

\subsection{Dataset Size and Sampling}

Each artifact yields 32 axial slices for mask prediction, resulting in 512 training slices per epoch.
For diffusion-based inpainting, one independently damaged volume is generated per artifact at each epoch.
Consequently, each training epoch consists of 16 volumetric samples, and over 100 epochs the diffusion model is trained on 1{,}600 distinct damage realizations.

%--------------------------------------
% Experiments, Results, and Discussion
%--------------------------------------
\section{Experiments, Results, and Discussion}
\label{sec:experiments}

We evaluate the proposed two-stage pipeline on held-out artifacts, measuring both geometric completion accuracy and color fidelity.
We first summarize the training protocol and hyperparameters, then define evaluation metrics, and finally present quantitative and qualitative results.

\subsection{Training Protocol}
\label{sec:training}

\paragraph{Stage 1: Damage Mask Prediction.}
UNet2DColor is trained to predict damage masks on $32\times32$ RGB slices extracted from voxelized artifacts.
We train for 50 epochs using Adam with learning rate $1\times10^{-4}$ and $(\beta_1,\beta_2)=(0.9,0.999)$, using a batch size of 8 slices.
The loss is binary cross-entropy.
Augmentations include random horizontal/vertical flips and in-plane rotations up to $\pm15^\circ$.
We select the checkpoint with the lowest validation loss.

\paragraph{Stage 2: Diffusion-Based Voxel Inpainting.}
VoxelInpaintUNet is trained to jointly reconstruct occupancy and color at $32^3$ resolution via a diffusion process.
We train for 100 epochs using Adam with initial learning rate $1\times10^{-3}$ and a \texttt{ReduceLROnPlateau} scheduler (factor 0.5, patience 5).
The batch size is 4 volumes with input shape $5\times32\times32\times32$.
We use a linear $\beta$ schedule from $10^{-4}$ to $2\times10^{-2}$ over $T=1000$ timesteps, sampling a random timestep each iteration.
Data augmentation includes mirroring along the axial direction with probability 0.5.
Model selection is based on validation loss and geometric metrics.

All experiments are run on a single NVIDIA A100 GPU.

\subsection{Evaluation Metrics}
\label{sec:metrics}

\paragraph{Geometry.}
We compare predicted occupancy $\widehat{V}$ against ground truth $V^{\mathrm{gt}}$ using Chamfer distance and F-score.
Chamfer distance is computed between occupied-voxel sets using nearest-neighbor queries (implemented with \texttt{scipy.spatial.cKDTree}).
Distances are reported in millimeters under the assumption that one voxel corresponds to one millimeter after normalization.
The F-score is computed at a 1~mm threshold as the harmonic mean of precision and recall over occupied voxels.

\paragraph{Color.}
Color fidelity is evaluated only on voxels occupied in both $\widehat{V}$ and $V^{\mathrm{gt}}$.
We report masked MSE and PSNR, and additionally compute PSNR per axial slice to analyze spatial variation.

\subsection{Quantitative Results}
\label{sec:quant}

\paragraph{Geometry completion.}
Table~\ref{tab:geom_results} compares Chamfer distance and F-score (1~mm) between the diffusion model and a symmetry-only baseline, where missing voxels are filled by mirroring occupied voxels across the dominant axial symmetry plane.
The diffusion model consistently improves both metrics, achieving the best performance around epoch 80.

\begin{table}[t]
  \centering
  \begin{tabular}{c|cc|cc}
    \hline
    Epoch & \multicolumn{2}{c|}{Diffusion Model} & \multicolumn{2}{c}{Symmetry Baseline} \\
          & Chamfer $\downarrow$ & F-score $\uparrow$ & Chamfer $\downarrow$ & F-score $\uparrow$ \\
    \hline
    10  & 0.0052 & 0.762 & 0.0123 & 0.512 \\
    50  & 0.0039 & 0.815 & 0.0110 & 0.533 \\
    80  & 0.0031 & 0.846 & 0.0106 & 0.545 \\
    100 & 0.0032 & 0.842 & 0.0105 & 0.548 \\
    \hline
  \end{tabular}
  \caption{Validation geometry metrics for the diffusion model versus a symmetry-only baseline. Lower Chamfer and higher F-score are better.}
  \label{tab:geom_results}
\end{table}

\paragraph{Color reconstruction.}
Table~\ref{tab:color_results} reports masked MSE and PSNR on overlapping occupied voxels.
Performance improves steadily through training and peaks around epoch 80.

\begin{table}[t]
  \centering
  \begin{tabular}{c|cc}
    \hline
    Epoch & Masked MSE $\downarrow$ & PSNR (dB) $\uparrow$ \\
    \hline
    10  & 0.00345 & 24.62 \\
    50  & 0.00221 & 26.56 \\
    80  & 0.00198 & 27.03 \\
    100 & 0.00205 & 26.89 \\
    \hline
  \end{tabular}
  \caption{Validation color metrics on overlapping occupied voxels. Lower MSE and higher PSNR are better.}
  \label{tab:color_results}
\end{table}

\subsection{Qualitative Results}
\label{sec:qual}

\begin{figure}[t]
  \centering
  \includegraphics[width=\linewidth]{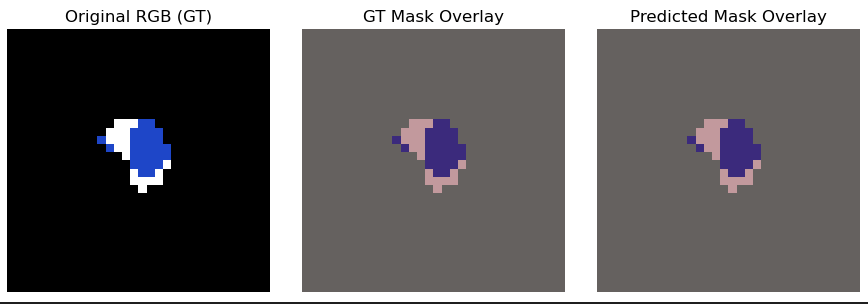}
  \caption{Stage 1 mask prediction: (left) RGB slice, (center) ground-truth mask overlay, (right) predicted mask overlay.}
  \label{fig:mask_example}
\end{figure}

\paragraph{Mask prediction.}
Figure~\ref{fig:mask_example} shows representative 2D slices with ground-truth and predicted damage masks overlaid on RGB.
The model captures irregular hole boundaries and local missing regions.

\begin{figure}[t]
  \centering
  \includegraphics[width=\linewidth]{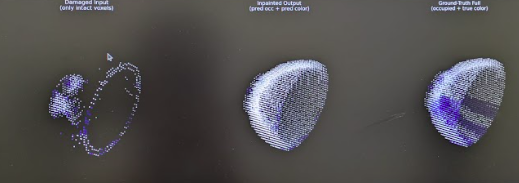}
  \caption{3D inpainting for an artifact with a distinct pattern: damaged input versus diffusion reconstruction and ground truth (as rendered in the figure).}
  \label{fig:qualitative_patterned}
\end{figure}

\begin{figure}[t]
  \centering
  \includegraphics[width=\linewidth]{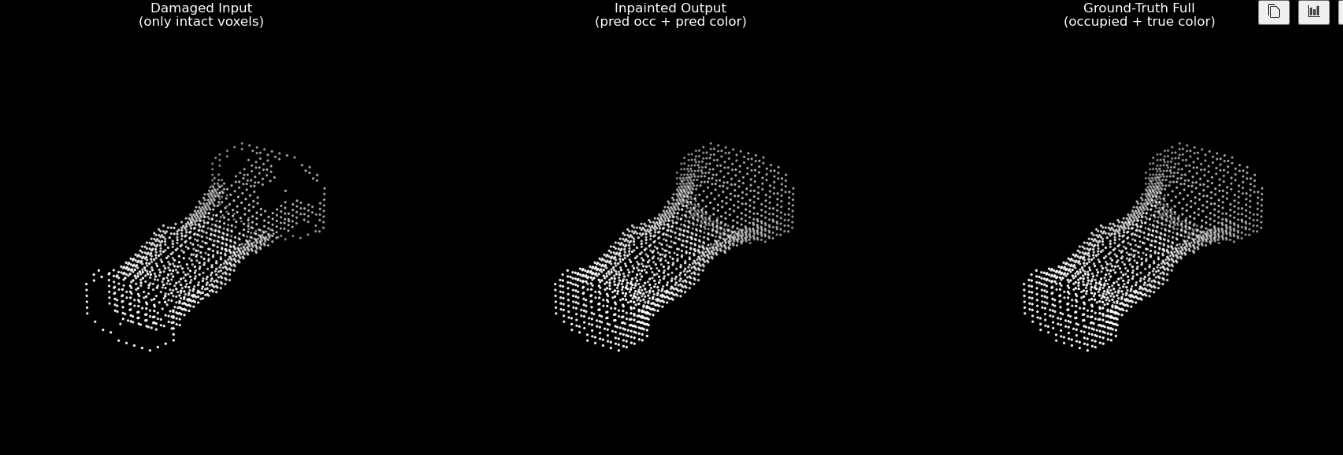}
  \caption{3D inpainting for an artifact with a nearly uniform surface: damaged input versus diffusion reconstruction and ground truth (as rendered in the figure).}
  \label{fig:qualitative_simple}
\end{figure}

\paragraph{3D inpainting.}
Figures~\ref{fig:qualitative_patterned} and~\ref{fig:qualitative_simple} show two representative reconstructions.
The diffusion model recovers missing structures and produces coherent color transitions relative to the damaged inputs.

\begin{figure}[t]
  \centering
  \includegraphics[width=\linewidth]{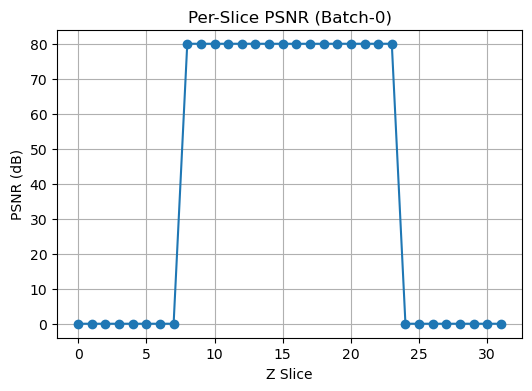}
  \caption{Per-slice PSNR on the validation set at epoch 80.}
  \label{fig:psnr_slices}
\end{figure}

\paragraph{Per-slice PSNR.}
Figure~\ref{fig:psnr_slices} plots PSNR across axial slices for the best-performing checkpoint (epoch 80).
Most slices exceed 25~dB, with dips in regions containing fine decorative patterns.

\subsection{Discussion and Limitations}
\label{sec:discussion}

Across held-out artifacts, mask conditioning combined with diffusion-based voxel inpainting yields substantial improvements over the symmetry-only baseline.
Geometry completion is strong, with missing structures recovered reliably under the $32^3$ voxelization.
Color inpainting is generally plausible but remains more sensitive to texture complexity and limited training diversity.
Artifacts with highly detailed or asymmetric surface patterns are the most challenging cases, where reconstructed colors can deviate from the ground truth despite correct geometry.

Future work could improve color fidelity by incorporating stronger appearance priors, higher-resolution representations, or multi-view image conditioning.

%--------------------------------------
% Conclusion and Future Work
%--------------------------------------
\section{Conclusion and Future Work}

This work introduces a lightweight framework for joint geometry and color inpainting in voxelized 3D objects, motivated by the digital restoration of damaged cultural heritage artifacts. By decoupling damage localization from reconstruction, the proposed pipeline allows the diffusion-based model to focus on filling missing regions while preserving existing structure. Experiments on a small but diverse set of textured artifacts demonstrate that explicit mask conditioning substantially improves geometric completion over symmetry-based baselines at the $32^3$ resolution used in our experiments.

While geometric reconstruction is consistently accurate at the evaluated resolution, color inpainting remains more challenging, particularly for artifacts with asymmetric or highly detailed surface patterns. These limitations highlight the difficulty of learning appearance priors from limited data and low-resolution volumetric representations.

Several directions could extend this work. Higher-resolution voxel grids or hybrid voxel–implicit representations may improve geometric and color fidelity. Incorporating stronger appearance cues, such as multi-view photographs, surface normals, or learned texture priors, could further stabilize color reconstruction. Finally, interactive or user-guided inpainting mechanisms may provide a practical balance between automation and expert control in real-world restoration workflows.
 
%--------------------------------------
% References
%--------------------------------------
\bibliographystyle{plainnat}
\bibliography{references}

\end{document}